# Learning Modular Simulations for Homogeneous Systems

**Jayesh K. Gupta**[1,*], **Sai Vemprala**[1,*], **and Ashish Kapoor**[1]

[*]Equal contributions, [1]Microsoft Autonomous Systems and Robotics Research

Complex systems are often decomposed into modular subsystems for engineering tractability. Although various equation based white-box modeling techniques make use of such structure, learning based methods have yet to incorporate these ideas broadly. We present a modular simulation framework for modeling *homogeneous* multibody dynamical systems, which combines ideas from graph neural networks and neural differential equations. We learn to model the individual dynamical subsystem as a neural ODE module. Full simulation of the composite system is orchestrated via spatio-temporal message passing between these modules. An arbitrary number of modules can be combined to simulate systems of a wide variety of coupling topologies. We evaluate our framework on a variety of systems and show that message passing allows coordination between multiple modules over time for accurate predictions and in certain cases, enables zero-shot generalization to new system configurations. Furthermore, we show that our models can be transferred to new system configurations with lower data requirement and training effort, compared to those trained from scratch.

## 1. Introduction

Modeling and simulation are key enablers in engineering as viable alternatives to real world systems. Simulation allows for the testing and evaluation of large/complex systems in a risk-free environment and has been of use in many disciplines. Simulation also plays a key role in domains such as deep reinforcement learning, which requires a large volume of interactions with the environment to learn effective policies. However, traditional simulators often require a large amount of engineering effort to create and can be computationally intensive when high fidelity predictions are needed. This has created interest in data-driven simulations, where machine learning algorithms are used to learn system dynamics directly from observations (Grzeszczuk et al., 1998; Markovsky et al., 2008; Hu, 2015; Kutz et al., 2016; Brunton et al., 2019). While machine learning algorithms excel at learning correlations from data, such end-to-end learning approaches result in models that cannot be reused for different system configurations. Additionally, end-to-end learning also requires large amounts of data and training effort when trying to model real-world systems.

In this paper we propose a machine learning based framework to learn and compose modular simulators. We build upon the observations that complex systems comprises of multiple modules (e.g., multibody systems), where each module evolves according to its own set of dynamical equations with corresponding inputs and outputs (Kübler et al., 2000). Examples include multiple pendulums connected via a single thread and multiple quadrotors carrying a large structure or a beam. Instead of modeling the entire system jointly, it might be more beneficial to model individual subsystems (such as a single pendulum or one quadrotor) and then consider interaction effects. The ability to harness such modularity can help reduce the sample complexity for machine learning as well create transferable subsystems that enable rapid prototyping.

The technical methodology we describe here builds upon a class of models referred to as *mechanistic models*. Mechanistic models aim to introduce inductive biases related to the underlying mechanisms of the systems being modeled, which are in turn informed by domain knowledge. Some examples are neural networks motivated by differential equations (Chen et al., 2018; Dupont et al., 2019; Tzen et al., 2019; Jia et al., 2019; Li et al., 2020; Liu et al., 2019; Holl et al., 2020), physics-informed neural networks (Raissi et al., 2019; Lutter et al., 2019; Cranmer et al., 2020; Gupta, Menda, Manchester, and M. J. Kochenderfer, 2019; Gupta, Menda, Manchester, and M. Kochenderfer, 2020; Misyris et al., 2020; Finzi et al., 2020) etc.

As mechanistic models encode structural knowledge of the system at hand, they can be used to learn modular simulations, as opposed to treating the entire system as a singular block.

In this work, we propose an approach to simulate homogeneous systems from data. Under the assumption that the systems of interest can be modeled as ordinary differential equations, we present a class of neural ODEs which we call *message-passing neural ODEs* (MP-NODE). Instead of modeling the entire graph of a system through ODEs, the message passing neural ODE only learns the model of the subsystem that forms represents a node in the graph. In order to enable interaction between the nodes, the MP-NODE contains additional message variables along with the state variables. The entire system is thus composed of several MP-NODE modules with the weights shared between them, and the coordination as well as ability to account for model approximation errors over time is learnt through the messages. We show that MP-NODEs allow for expressive and efficient modular simulations which can be fine-tuned to new system configurations significantly faster than training a model from scratch. Our key contributions are listed below.

1. We present the message-passing neural ODE (MP-NODE), an augmented neural ODE formulation aimed at enabling interaction between multiple components through messages.
2. We test the MP-NODE on various connected systems and demonstrate its ability to learn complex dynamics and generalize to unseen system configurations.
3. We also show that this modular framework allows finetuning for new configurations with significantly lower training effort than training models from scratch.

## 2. Related Work

Let us assume $X_t \in \mathcal{X}$ is the state of the world at time $t$. Then, dynamically stepping over $T$ timesteps with optional control input $U_t \in \mathcal{U}$ gives a trajectory of states $\mathbf{X}_{t_{0:T}} = (X_{t_0}, \ldots, X_{t_T})$. A *simulator* $s : \mathcal{X} \times \mathcal{U} \to \mathcal{X}$, models the dynamics by mapping previous states and control inputs to future states. Traditional simulators often define the dynamics as a modular set of differential or differentiable algebraic equations and use appropriate numerical solvers as the update mechanism.

For data-driven simulations, we parameterize the dynamics of the simulator as $f_\theta : \mathcal{X} \times \mathcal{U} \to \mathcal{Y}$. Here, the parameters $\theta$ can be optimized for some training objective. The semantics of the dynamics information $Y \in \mathcal{Y}$ is governed by an update function, usually an integrator.

Learning such simulations from data has been quite important in fields like physics (He et al., 2019) and graphics (Ladick et al., 2015). Learned simulations can often be faster and more efficient for complex phenomenon prediction than just equation based simulators (De Bézenac et al., 2019; Kochkov et al., 2021).

Combining graph neural networks (GNN) (Scarselli et al., 2008; Battaglia et al., 2016; Kipf et al., 2016) and differential equations (Chen et al., 2018; Dupont et al., 2019) has been explored in Poli et al. (2019), Huang et al. (2021) and Zang et al. (2020). The dynamics of their test systems are relatively simple. Importantly, there is no study of transferability of the learned dynamics to changing graph structure. More recently, graph neural networks have been applied in the context of particle simulations by Sanchez-Gonzalez et al. (2020) and mesh-based simulations by Pfaff et al. (2020). Moreover, unlike other graph neural net based dynamics learning works (Poli et al., 2019; Zang et al., 2020) they also use edge based features. However, these largely ignore the existence of control inputs, limiting their usefulness for reinforcement learning and control applications.

Some work in the past has examined the idea of message passing for continuous time Bayesian networks (El-Hay et al., 2010), or as a combination of generative graphical models and GNNs (Satorras et al., 2019). The focus of these works was to model full system dynamics for a particular graph structure, where messages play a role in computing the interactions, but do not evaluate transfer performance to different graph structures. We distinguish our work from such prior methods primarily through the fact that our method is generalizable between different graph structures. An approach that is close to ours from the neural ODE body of work would be the augmented neural ODEs (Dupont et al., 2019) which also contain extra dimensions in the state, although these additional variables are zero-valued. In our method, we repurpose those extra dimensions to explicitly pass messages across multiple neural ODEs as per the graph structure.

## 3. Methodology

The proposed framework tackles the problem of learning modular simulations where (1) smaller subsystems are individually modeled and (2) can be composed to simulate larger systems and processes. Specifically, given observations from a complex system, the methodology seeks to learn both the individual simulation module as well the message passing mechanism that would best explain the data traces.

At the core of this method are Neural Differential Equations that use messages to coordinate amongst themselves. Such message passing enables the framework to model various interaction effects between the modules as well as account for errors due to model approximations and time discretization. The next subsection reviews Neural ODEs. We then present our MP-NODE which extends neural ODEs to incorporate message passing. Note that we assume that the graphs are homogeneous and the underlying graph structure is given. However, we do show that the learnt Neural ODEs can be reused across different given graph topologies. Handling heterogeneous subsystems along with learning the underlying graph structure is a more challenging problem and will be the focus of future work.

### 3.1. Background: Neural Differential Equations

Many continuous-time dynamical systems which can be found in physics (oscillators, mechanical systems), biology (population dynamics) etc. are best modeled by ordinary differential equations. Typically, such a system can be described using an ODE as $\dot{x} = f(x, u, t)$. A data-driven way of learning a simulation of this system would thus be to estimate $\theta$ such that $\dot{x} = f_\theta(x, u, t)$ fits a set of given observations.

Neural ordinary differential equations (Chen et al., 2018) are a class of neural networks that link the concepts of residual networks and dynamical systems. Given a set of observations of the system state over time, under the assumption that the system dynamics can be described through an ODE, neural ODEs can be used to approximate the time derivative directly. By integrating the model output using a black-box ODE solver for an arbitrarily long timestep and backpropagating appropriately, the model parameters $\theta$ can be updated to minimize the distance between the predicted states (Equation 2) and the observed ones.

$$\frac{d\hat{\mathbf{x}}(t)}{dt} = \mathbf{f}_\theta\left(\mathbf{x}(t), \mathbf{u}(t), t\right) \tag{1}$$

$$\hat{\mathbf{x}}(t+1) = \mathbf{x}(t) + \int_t^{t+1} \mathbf{f}_\theta\left(\mathbf{x}(t), \mathbf{u}(t), t\right) dt \tag{2}$$

### 3.2. Message Passing with Neural ODEs (MP-NODE)

Now, let us consider a connected system that evolves as $\frac{dX}{dt} = f(G, X, U, t)$, where $X(t) \in \mathbb{R}^{n \times d}$ represents the entire system state, constituting of $n$ connected nodes whose state is of length $d$ each. $G = \{\mathcal{V}, \mathcal{E}\}$ captures the graph structure and how the nodes are connected to each other, which we assume is known. $\mathcal{N}(k)$ refers to the neighbors of a node $k$ in the graph.

While neural ODEs have been successfully used to model dynamical systems, modeling large systems with complex connectivity and dynamics is still challenging and computationally intensive. Furthermore, even the slightest change to the connectivity of the system necessitates retraining of the entire system dynamics model. On the other hand, connected systems can often be decomposed into repeatable subsystems that communicate with each other, hence it can be more efficient to learn these individual blocks rather than attempting to model the connected system as a whole.

The first key aspect of our framework is that the focus is placed on the subsystems rather than the entire system. For a given homogeneous network, we essentially only learn the model corresponding to a single subsystem, or node. When orchestrating full simulation of the system, the weights are shared across all individual nodes.

Secondly, due to this modular nature, the individual nodes need to learn to coordinate with each other and understand how their long term evolution changes based on interaction according to the graph structure. We facilitate such communication between the nodes by augmenting each subsystem state with additional variables which we call *messages* with the augmented state for node $k$ being $\mathbf{X}^k = \begin{bmatrix} \mathbf{x}^k \\ \mathbf{m}^k \end{bmatrix}$. With the neural ODE operating on the augmented state, each neural subsystem outputs a message along with its predicted state at each timestep. Similarly, each node also receives a message at the input, which results from an aggregation operation between all output messages of the node neighbors which are identified according to $G$. We call this model the *message passing neural ODE* (MP-NODE) and describe the structure in Figure 1(Left). The full formulation of the MP-NODE can thus be written as below, where $\mathbf{m}'_k(t)$ is the input message at node $k$ and $\mathbf{m}_k$ is the output message.

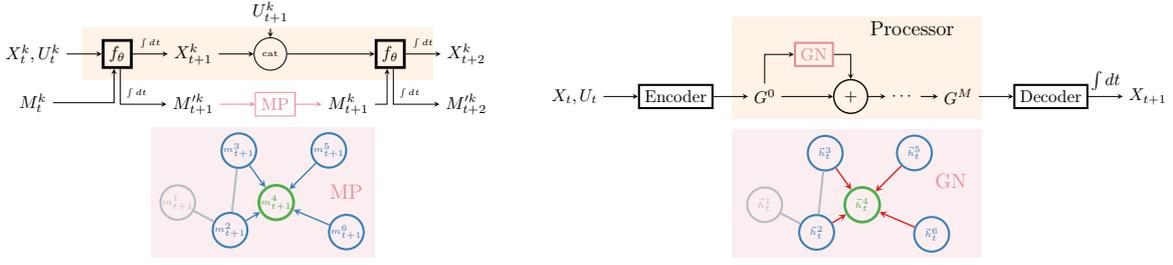

**Figure 1:** (Left) MP-NODE architecture: Individual nodes can simulate independently and coordinate via message passing. (Right) Learning to Simulate (L2S) architecture without edge features.

$$\mathbf{m}'_k(t) = \mathrm{MEAN}\left(\{\mathbf{m}_j(t), j \in \mathcal{N}(k)\}\right) \tag{3}$$

$$\frac{d}{dt}\begin{bmatrix}\mathbf{x}_k(t)\\\mathbf{m}_k(t)\end{bmatrix} = \mathbf{f}_\theta\left(\begin{bmatrix}\mathbf{x}_k(t)\\\mathbf{m}'_k(t)\end{bmatrix}, \mathbf{u}(t)\right) \tag{4}$$

$$\hat{\mathbf{X}}_k(t+dt) = \mathrm{ODESOLVE}\left(\mathbf{f}_\theta\left(\begin{bmatrix}\mathbf{x}_k(t)\\\mathbf{m}'_k(t)\end{bmatrix}, \mathbf{u}(t), dt\right)\right) \tag{5}$$

While the outputs from the ODE solver contain both the predicted state and a message, the supervision and loss computation only happen on the predicted states, allowing the messages to evolve freely. Over time, the model is able to use these additional variables to potentially encode the effects of different aggregations and interactions between neighbors, finally reaching an optimum configuration that is applicable for all the nodes. Learning such a generalizable subsystem allows the MP-NODE framework to be directly applicable for another system configuration with higher or lower number of nodes, as opposed to attempting to learn dynamics for the entire system state such as is done with a conventional neural ODE, which needs retraining for new configurations.

The length of the message can be varied, and we treat it as a hyperparameter that changes the expressivity of the model. In our implementation, we use the mean as the aggregator for incoming messages from the neighbors of a node.

## 4. Experiments

We evaluate the performance of MP-NODE on five different systems: two of them selected to focus on robotic regimes and three of them as representative of standard graph neural network benchmarks, for all of which we generate datasets of trajectories. We implement our method in Julia (Bezanson et al., 2017) and make use of the SciML ecosystem (Christopher Rackauckas et al., 2017; Innes et al., 2018). We list training-specific details such as hyperparameters, learning rates, optimizers in the appendix. Source code for our experiments is available at https://github.com/microsoft/MPNODE.jl.

We list the systems used below, and provide additional details in the appendix.

- *Coupled Pendulum:* This system is a combination of two identical simple pendula connected by a string, behaving in an undamped manner. Hence, we train an MP-NODE that models a single pendulum, which is instantiated twice and the communication is enabled through the messages.
- *Lorenz Attractor:* A Lorenz system is a simplified model of atmospheric convection, described through a 3-dimensional system that exhibits chaotic behavior in specific cases Lorenz, 1963. In our case, we consider multiple Lorenz systems coupled in a fully connected manner. We denote Lorentz3 and Lorentz10 as fully connected systems with 3 and 10 nodes respectively.
- *Gene Dynamics:* This system simulates a continuous time system from biology: the gene regulatory dynamics governed by the Michaelis-Menten equation(Alon, 2019) over grid-like configurations with

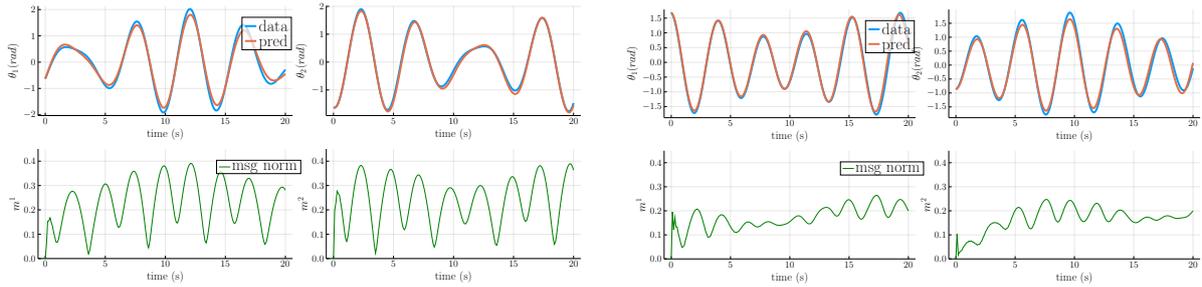

**Figure 2:** The figure compares the predicted (red) with observed (blue) state evolution of the coupled pendulum system. In the first row, the two pairs of columns are two different instance of trajectories. Each column represent one pendulum and the green curves represents outgoing messages. The messages enable interaction between the connected pendulums and enable accurate modeling.

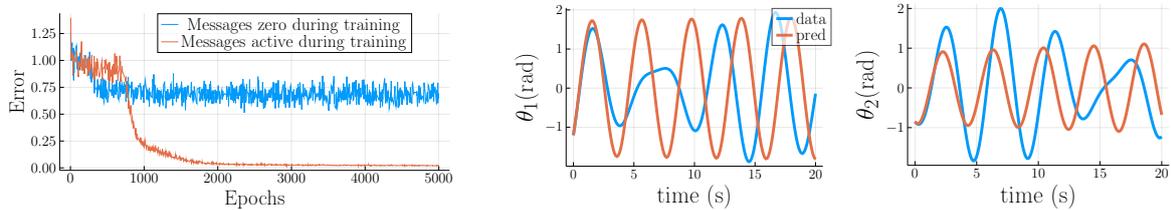

**(a)** Comparison of normal messages vs. set to zero during training. Training loss is much lower when messages are active (red) vs. when they are set to zero (blue).

**(b)** When message passing is disabled at test time, each node acts similar to an unconnected, simple pendulum. The two graphs represent the two nodes and we plot the prediction (red) and the observed (blue) state evolution.

**Figure 3:** MP-NODE applied to the coupled pendulum system. Together with Figure 2 we show how the messages help minimize drift and compare the characteristics of the system when messages are disabled.

varying connectivity. Similar to one of the test cases in (Zang et al., 2020), we generate data for three types of graph topologies: a) Erdós-Rényi (ER) (Erds et al., 1960); b) Barabási-Albert (BA) (Barabási et al., 1999) and c) Wattz-Strogatz (WS) (Watts et al., 1998). We use two configurations in our experiments, where the dynamics are simulated over a $4 \times 4$ grid and an $8 \times 8$ grid.

- *Kuramoto System:* The Kuramoto model (Kuramoto, 2003) describes the behavior of large sets of coupled oscillators. We simulate the Kuramoto dynamics over the same three types of networks mentioned above, for 10 nodes, which we refer to as Kuramoto10. We use Lindner et al. (2020) to build our reference simulators to generate our datasets.
- *Quadrotor swarm:* This system involves a more challenging dynamics learning task of multiple quadrotors carrying a heavy load (that a single quadrotor cannot lift) while passing through a narrow doorway (Jackson et al., 2020). We consider two versions of this: one with 3 quadrotors, another with 6.

We compare our proposed approach to the one proposed by Sanchez-Gonzalez et al. (2020) that generalized different graph based dynamics learning methods, which we refer to as the L2S baseline. We do not use edge features in our implementation of L2S (Figure 1(Right)).

### 4.1. Effect of messages

When the MP-NODE method of modeling is applied to systems with connected nodes, we find that over time, the messages evolve to enable coordination between the nodes. Figure 2 compares predictions (red) with observations (blue) of state evolution for the coupled pendulum system. Each column corresponds to a pendulum and we plot two different trajectories (rows 1 and 3) in Figure 2. We observe that the predictions closely match the ground truth observations. We also plot the norm of outgoing messages (green) (rows 2 and 4) in Figure 2 and notice that they dynamically change in order to appropriately coordinate across the two nodes. We also investigate the importance of message passing by comparing performance of MP-NODE with and without messages. Specifically, we

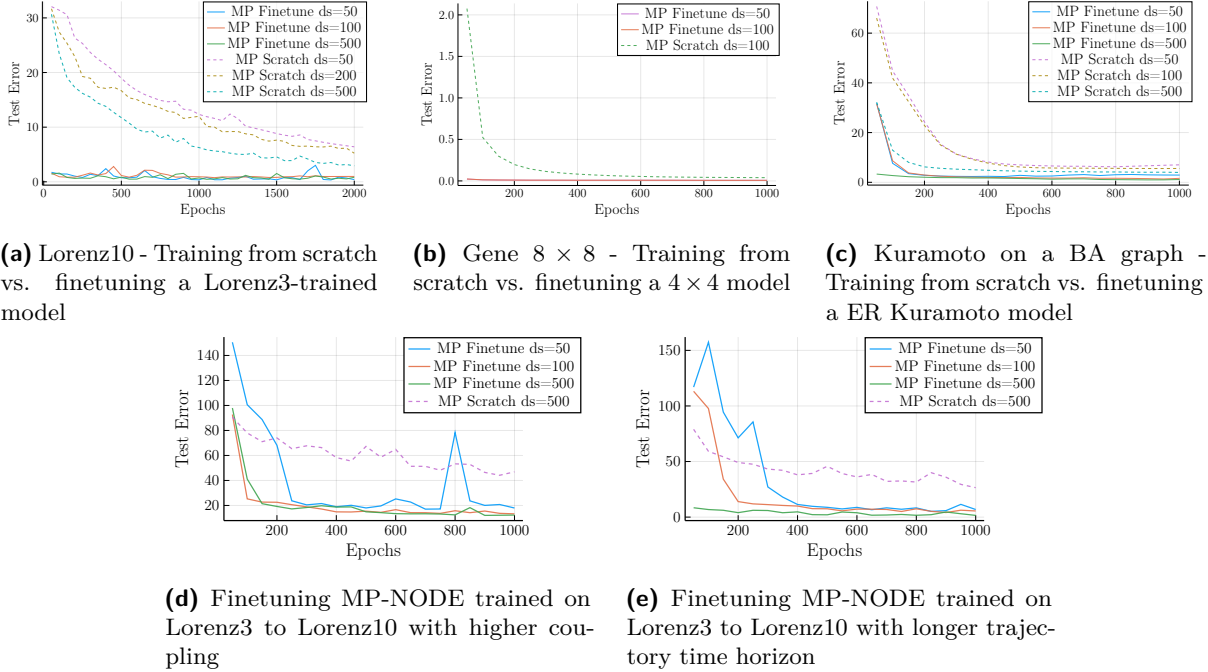

**(a)** Lorenz10 - Training from scratch vs. finetuning a Lorenz3-trained model

**(b)** Gene 8 × 8 - Training from scratch vs. finetuning a 4 × 4 model

**(c)** Kuramoto on a BA graph - Training from scratch vs. finetuning a ER Kuramoto model

**(d)** Finetuning MP-NODE trained on Lorenz3 to Lorenz10 with higher coupling

**(e)** Finetuning MP-NODE trained on Lorenz3 to Lorenz10 with longer trajectory time horizon turn the messages off by clamping the message outputs to zero. Note that such forcing the messages to zero is similar to the augmented ODE applied to individual nodes. Figure 3a shows that we achieve lower training loss with messages on (red) vs when off (blue). We again test the idea of forcing messages to zero during training (similar to augmented neural ODE) on the Kuramoto system.

**Figure 5:** MP-NODE performance when trained from scratch vs. finetuning an existing model

As can be seen from Figure 4, using the additional dimensions for message-passing between nodes as in MP-NODE leads to substantial improvements in learning performance. This is consistent with the findings for the coupled pendulum system.

Similarly, Figure 3b highlights that if no messages are passed between the nodes at test time, each node starts behaving like a simple pendulum that is unconnected to any neighbors. This indicates that the message parameters are important and help the model by encoding the interaction between the nodes.

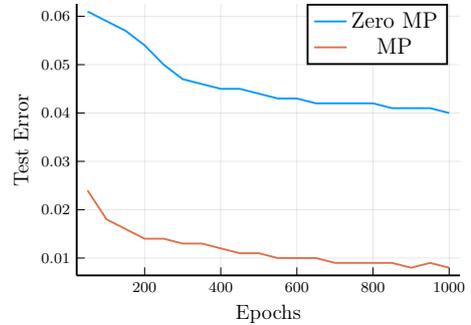

**Figure 4:** Performance on Kuramoto System: Normal MP-NODE vs. MP-NODE with messages set to zero

### 4.2. Finetuning

The modular nature of the MP-NODE also allows for easy transfer to different configurations of systems. We examine the ability and performance of finetuning in several cases.

**Increasing number of nodes:** A common use case would be to take a trained MP-NODE module, representing a subsystem, and finetune it for a larger graph than what it was originally trained on. We use the MP-NODE that was trained on Lorenz3, and finetune this model for the Lorenz10 configuration. We compare this training effort with that of MP-NODE models trained from scratch just for Lorenz10. We show in Figure 5a that even with a small dataset of 50 trajectories, finetuning a pretrained MP-NODE reaches a low error significantly faster than the models trained from scratch with larger training sets.

Similarly, we put the modular nature of the MP-NODE to test in the gene dynamics system by taking MP-NODE models trained with smaller grid sizes and finetuning them on data from larger grid sizes. In Figure 5b, we show the finetuning performance when a model trained on a 4 × 4 grid is transferred to an

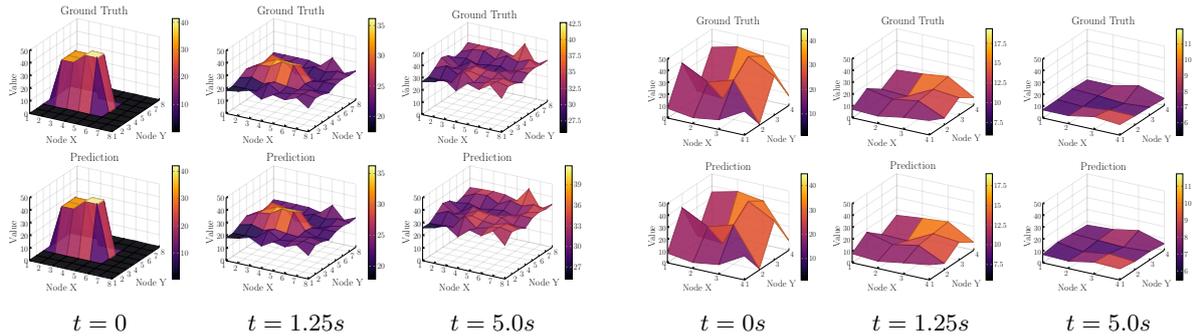

**(a)** MP-NODE predictions on a 8 × 8 grid of gene regulatory dynamics finetuned over a model trained on 4 × 4 grid data.

**(b)** MP-NODE predictions on a 4 × 4 grid of gene regulatory dynamics using the ER network, a topology unseen during training.

**Figure 6:** Performance of the MP-NODE on the gene dynamics model: finetuning to larger grids, as well as zero-shot generalization to unseen graph structures.

**Table 1:** Zero-shot generalization performance of MP-NODE

| Validation | | Test | | |
| --- | --- | --- | --- | --- |
| BA | BA - L2S | BA | ER (unseen) | WS (unseen) |
| $0.0051 \pm 0.0001$ | 0.0106 | $0.0045 \pm 0.0004$ | $0.0026 \pm 0.0008$ | $0.0023 \pm 0.0002$ |

8 × 8 grid, compared to one trained from scratch on the 8 × 8 grid. Similar to the trend with the Lorenz systems, we see that finetuning a trained MP-NODE allows for accurate predictions faster than training a model from scratch. Furthermore, in Figure 6a, we show a qualitative comparison of ground truth and predicted state over the finetuned Gene 8 × 8 grid at different time steps.

This shows that due to the modular characteristic of our approach, finetuning a pretrained MP-NODE module allows for rapid adaptation to arbitrary number of nodes in a graph.

**Higher system complexity:** We also evaluate the possibility of finetuning MP-NODEs for different system parameters. As an example, we finetune a MP-NODE model trained on Lorenz3 to Lorenz10 but with different coupling intensity and show the results in Figure 5d. We observe that finetuning requires a lot less data to achieve better performance than training from scratch. Similarly, when we finetune the MP-NODE model trained on Lorenz3 to Lorenz10 for longer time horizon of 10s in Figure 5e, we again find that a lot less data is required to achieve better performance than training from scratch.

**Changing graph structure:** We also examine the ability of MP-NODE to be finetuned for different graph structures. To this end, we train an MP-NODE on a Kuramoto10 system connected according to the Barabasi-Albert (BA) network and attempt to finetune it for systems of other network types. Similar to above, we see that finetuning is more efficient at adapting to new network types than training models from scratch. A comparison of training effort is shown in Figure 5c when transferring the Kuramoto10 model trained on the BA network to the ER network.

### 4.3. Zero-shot generalization

Considering that the modular nature of MP-NODE allows us to connect arbitrary number of subsystems together for a given graph structure, we can investigate the performance for new configurations without explicitly finetuning the model. When tested on the Lorenz system, we find that an MP-NODE trained only on Lorenz3 exhibits reasonable zero-shot generalization performance on a higher number of Lorenz attractors, such as Lorenz7 and Lorenz10 (see appendix) without requiring any additional training.

We notice that this ability holds for not only changing numbers of nodes, but also different graph topologies. We train the MP-NODE on the Gene 4 × 4 system connected according to the Barabasi-Albert (BA) topology, with the dataset comprising of data corresponding to five different adjacency matrices in the BA topology. By training on data coming from different types of connections, the MP-NODE learns how node dynamics are are affected by different graph structures. After training, we observe that the

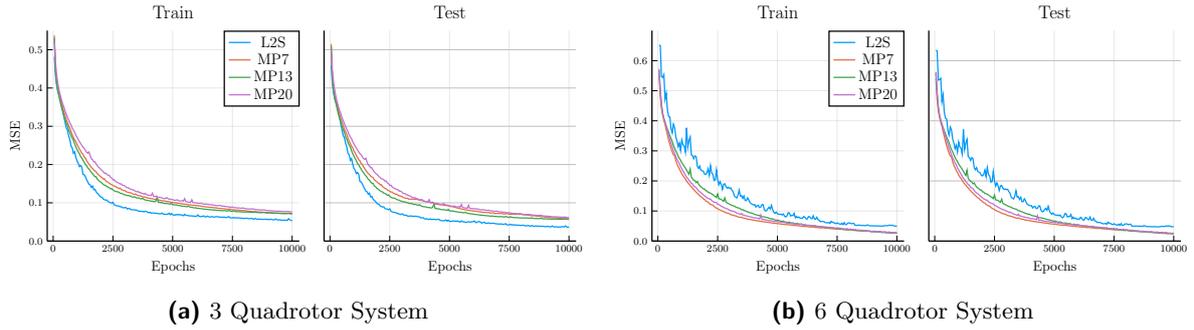

**(a)** 3 Quadrotor System

**(b)** 6 Quadrotor System

**Figure 7:** Effect of message dimensionality on train/test performance of the MP-NODE examined for Quadrotor systems

MP-NODE model that was trained only on BA topology generalizes to the other unseen topologies as well (ER and WS). Table 1 shows the MP-NODE's performance on both seen and unseen network topologies. We observe that MP-NODE exhibits low test error on unseen topologies as well, while achieving better validation performance than L2S. In Figure 6b, we take a qualitative look at the predictions for gene dynamics on the ER network using the zero-shot transferred model.

### 4.4. Effect of message size

In order to further improve the performance when modeling generally complex dynamics of systems such as Kuramoto or the quadrotor swarm, we investigate whether having a larger message size helps. We train MP-NODE models on the Kuramoto10 system with message dimensionalities of 1, 3, 7, and 13. By default, each Kuramoto node has a state length of 3. We show this comparison in Figure 8. We notice that larger messages result in higher accuracy, but after a certain threshold (messsage size = 7 in this case), the improvement becomes minimal.

In the case of the quadrotor swarm, we choose message sizes of 7, 13, 20 for analysis. Through Figure 7, we observe that all tested message sizes for MP-NODE lead to similar final test errors. Although L2S seems to do slightly better on the 3 Quadrotor System, it performs worse on the 6 Quadrotor System.

Similar to the trend in other datasets, we find in Figure 9 that finetuning MP-NODE on even limited data allows learning more accurate simulations much faster than learning from scratch, and this holds true for different message sizes as well. Although, we note that higher message dimensions can lead to higher data requirements.

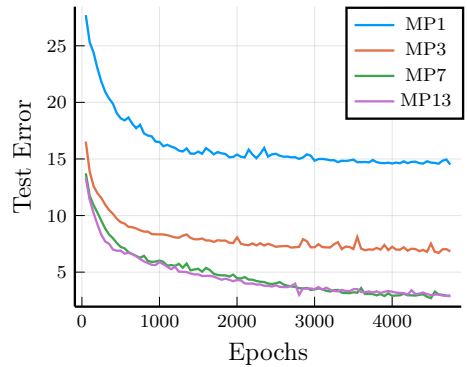

**Figure 8:** Effect of message size on test performance in Kuramoto system

### 4.5. Summary

To better demonstrate the advantages of MP-NODE when it comes to finetuning for different system configurations, we consider a metric: for a given test configuration, we compute the 'number of epochs taken to reach minimum test error multiplied by the value of the minimum test error observed'. Intuitively, we can see that a lower number is better. We summarize our findings in Table 2, where we compare MP-NODE finetuning vs. training from scratch, and for some configurations, we also compare with L2S. In the case of Lorenz and Kuramoto systems, we evaluate the performance under varying number of trajectories in the finetuning dataset. Through Figure 9, we observe that MP-NODE finetuning consistently achieves a lower combination of test error and number of epochs required to reach it, even in the low-data regime.

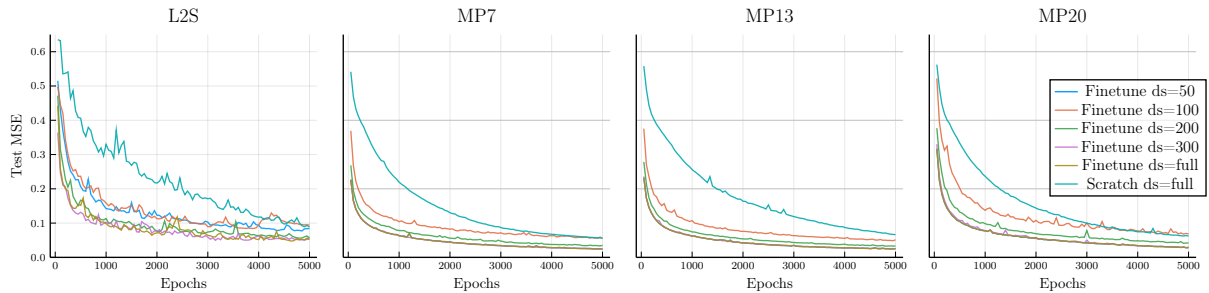

**Figure 9:** Performance of a 3 Quadrotor MP-NODE finetuned on the 6 Quadrotor System

**Table 2:** Summary of the performance of MP-NODE finetuning vs. other baselines, using the metric of minimum test error multiplied by number of epochs taken to reach the minimum. Lower is better.

|  | **Lorenz10** | | | **Kuramoto** | | | **Gene Dynamics** | **Quadrotors** |
| --- | --- | --- | --- | --- | --- | --- | --- | --- |
| Model | ds=50 | ds=100 | ds=500 | ds=50 | ds=100 | ds=500 | | |
| MP-NODE finetuned | **632.70** | **833.91** | **477.75** | **1358.50** | **1264.83** | **874.80** | **12.53** | **236.00** |
| MP-NODE scratch | 12800 | 10454 | 5656 | 5008 | 5268 | 3974 | 28 | N/A |
| L2S | N/A | N/A | N/A | N/A | N/A | N/A | 46.64 | 476.0 |

## 5. Conclusions

We present a framework for modular modeling of complex homogeneous systems called the message passing neural ODE (MP-NODE). MP-NODE builds upon neural ODEs, and learns a model for the individual node within a connected homogeneous system, which can then be shared across all instances in the system. Augmenting the system state with additional message variables allows the model to encode the features required for coordination and accurate prediction over long horizons. This modular way of thinking about complex systems opens up several possibilities such as zero-shot generalization to unseen configurations and rapid finetuning to more complex configurations. We apply our framework to a wide variety of systems such as coupled pendulum, Lorenz attractors, biological networks, Kuramoto systems and a swarm of quadrotors, and showcase its ability to learn complex dynamics in a modular fashion while outperforming existing methods. We also show that pretrained MP-NODE models can be finetuned for new configurations significantly faster than training from scratch which has the potential to greatly amortize data and compute requirements for data-driven simulations. While we focused on neural ODEs, general principles of parameter sharing and message passing as used in this work would be applicable to modern state-space models (Gu et al., 2021) too.

**Limitations:** Current implementation of MP-NODE is limited to homogeneous systems only. For better ability to adapt to more complex systems, an extension that can handle heterogeneous systems would be needed. Although encoder-decoder architectures could help in extending to nodes with different node state dimensions, different time-stepping requirements for the nodes might require a different messaging setup. Moreover, the approach currently requires full knowledge of the graph structure which may not always be available for large complex systems. In our current problem setup, we also do not consider potential issues such as delays in communication between nodes, which we leave for future work.

**Broader Impact:** Pretrained simulation nodes that are reusable can minimize data, compute, and energy requirements. Data-driven simulations can often speed up conventional slow simulations while also being capable of capturing real world phenomena ignored in traditional simulations. We show through our analysis of finetuning and generalization that our approach of modeling subsystems that are inherently reusable as opposed to specific configurations of systems has the potential to alleviate data and compute requirements. However, limited data and lack of physical constraints can lead to arbitrarily bad predictions for unseen simulation regimes. Nevertheless, building upon this approach, there is a potential to enable large variety of existing simulations to work with heterogeneous compute hardware and data via surrogate modeling (Chris Rackauckas et al., 2021).

# References

Alon, Uri (2019). *An introduction to systems biology: design principles of biological circuits*. CRC press.

Barabási, Albert-László and Réka Albert (1999). "Emergence of scaling in random networks." *science* 286.5439, pp. 509–512.

Battaglia, Peter W, Razvan Pascanu, Matthew Lai, Danilo Rezende, and Koray Kavukcuoglu (2016). "Interaction networks for learning about objects, relations and physics." *arXiv preprint arXiv:1612.00222*.

Bezanson, Jeff, Alan Edelman, Stefan Karpinski, and Viral B Shah (2017). "Julia: A fresh approach to numerical computing." *SIAM review* 59.1, pp. 65–98.

Brunton, Steven L. and J. Nathan Kutz (Jan. 2019). *Data-Driven Science and Engineering*. Cambridge University Press.

Chen, Ricky TQ, Yulia Rubanova, Jesse Bettencourt, and David Duvenaud (2018). "Neural ordinary differential equations." In: *Advances in Neural Information Processing Systems (NeurIPS)*.

Cranmer, Miles, Sam Greydanus, Stephan Hoyer, Peter Battaglia, David Spergel, and Shirley Ho (2020). "Lagrangian neural networks." *arXiv preprint arXiv:2003.04630*.

De Bézenac, Emmanuel, Arthur Pajot, and Patrick Gallinari (2019). "Deep learning for physical processes: Incorporating prior scientific knowledge." *Journal of Statistical Mechanics: Theory and Experiment* 2019.12, p. 124009.

Dörfler, Florian and Francesco Bullo (2014). "Synchronization in complex networks of phase oscillators: A survey." *Automatica* 50.6, pp. 1539–1564. ISSN: 0005-1098.

Dupont, Emilien, Arnaud Doucet, and Yee Whye Teh (2019). "Augmented Neural ODEs." In: *Advances in Neural Information Processing Systems (NeurIPS)*. Vol. 32.

Erds, P. and A Rényi (1960). "On the Evolution of Random Graphs." In: *Publication of the Mathematical Institute of the Hungarian Academy of Sciences*.

Finzi, Marc, Ke Alexander Wang, and Andrew Gordon Wilson (2020). "Simplifying hamiltonian and lagrangian neural networks via explicit constraints." *arXiv preprint arXiv:2010.13581*.

Grzeszczuk, Radek, Demetri Terzopoulos, and Geoffrey Hinton (1998). "Neuroanimator: Fast neural network emulation and control of physics-based models." In: *Proceedings of the 25th annual conference on Computer graphics and interactive techniques*.

Gu, Albert, Karan Goel, and Christopher Re (2021). "Efficiently Modeling Long Sequences with Structured State Spaces." In: *International Conference on Learning Representations (ICLR)*.

Guo, Yufeng, Dongrui Zhang, Zhuchun Li, Qi Wang, and Daren Yu (2021). "Overviews on the applications of the Kuramoto model in modern power system analysis." *International Journal of Electrical Power & Energy Systems* 129, p. 106804. ISSN: 0142-0615.

Gupta, Jayesh K, Kunal Menda, Zachary Manchester, and Mykel Kochenderfer (2020). "Structured mechanical models for robot learning and control." In: *Learning for Dynamics and Control*.

Gupta, Jayesh K, Kunal Menda, Zachary Manchester, and Mykel J Kochenderfer (2019). "A general framework for structured learning of mechanical systems." *arXiv preprint arXiv:1902.08705*.

El-Hay, Tal, Ido Cohn, Nir Friedman, and Raz Kupferman (2010). "Continuous-Time Belief Propagation." In: *ICML*.

He, Siyu, Yin Li, Yu Feng, Shirley Ho, Siamak Ravanbakhsh, Wei Chen, and Barnabás Póczos (2019). "Learning to predict the cosmological structure formation." *Proceedings of the National Academy of Sciences* 116.28, pp. 13825–13832.

Holl, Philipp, Vladlen Koltun, and Nils Thuerey (2020). "Learning to control PDEs with differentiable physics." *arXiv preprint arXiv:2001.07457*.

Hu, Xiaolin (2015). "Dynamic data-driven simulation: Connecting real-time data with simulation." In: *Concepts and Methodologies for Modeling and Simulation*. Springer, pp. 67–84.

Huang, Zijie, Yizhou Sun, and Wei Wang (2021). "Coupled Graph ODE for Learning Interacting System Dynamics." In: *Proceedings of the 27th ACM SIGKDD Conference on Knowledge Discovery & Data Mining*. New York, NY, USA: Association for Computing Machinery, pp. 705–715. ISBN: 9781450383325. URL: https://doi.org/10.1145/3447548.3467385.

Innes, Michael, Elliot Saba, Keno Fischer, Dhairya Gandhi, Marco Concetto Rudilosso, Neethu Mariya Joy, Tejan Karmali, Avik Pal, and Viral Shah (2018). "Fashionable modelling with flux." *arXiv preprint arXiv:1811.01457*.

Jackson, Brian E, Taylor A Howell, Kunal Shah, Mac Schwager, and Zachary Manchester (2020). "Scalable cooperative transport of cable-suspended loads with UAVs using distributed trajectory optimization." *IEEE Robotics and Automation Letters* 5.2, pp. 3368–3374.

Jia, Junteng and Austin R Benson (2019). "Neural jump stochastic differential equations." *arXiv preprint arXiv:1905.10403*.

Kingma, Diederik P and Jimmy Lei Ba (2015). "Adam: A method for stochastic gradient descent." In: *International Conference on Learning Representations (ICLR)*.

Kipf, Thomas N and Max Welling (2016). "Semi-supervised classification with graph convolutional networks." *arXiv preprint arXiv:1609.02907*.

Kochkov, Dmitrii, Jamie A Smith, Ayya Alieva, Qing Wang, Michael P Brenner, and Stephan Hoyer (2021). "Machine learning–accelerated computational fluid dynamics." *Proceedings of the National Academy of Sciences* 118.21.

Kübler, R and W Schiehlen (2000). "Modular simulation in multibody system dynamics." *Multibody System Dynamics* 4.2, pp. 107–127.

Kuramoto, Yoshiki (2003). *Chemical oscillations, waves, and turbulence.* Courier Corporation.

Kutz, J Nathan, Steven L Brunton, Bingni W Brunton, and Joshua L Proctor (2016). *Dynamic mode decomposition: data-driven modeling of complex systems.* SIAM.

Ladick, L'ubor, SoHyeon Jeong, Barbara Solenthaler, Marc Pollefeys, and Markus Gross (2015). "Data-driven fluid simulations using regression forests." *ACM Transactions on Graphics (TOG)* 34.6, pp. 1–9.

Li, Xuechen, Ting-Kam Leonard Wong, Ricky TQ Chen, and David Duvenaud (2020). "Scalable gradients for stochastic differential equations." In: *International Conference on Artificial Intelligence and Statistics (AISTATS)*.

Lindner, Michael, Lucas Lincoln, Fenja Drauschke, Julia Monika Koulen, Hans Würfel, Anton Plietzsch, and Frank Hellmann (2020). "NetworkDynamics.jl – Composing and simulating complex networks in Julia." *arXiv preprint arXiv:2012.12696*.

Liu, Xuanqing, Si Si, Qin Cao, Sanjiv Kumar, and Cho-Jui Hsieh (2019). "Neural SDE: Stabilizing neural ode networks with stochastic noise." *arXiv preprint arXiv:1906.02355*.

Lorenz, Edward N (1963). "Deterministic nonperiodic flow." *Journal of atmospheric sciences* 20.2, pp. 130–141.

Lutter, M, C Ritter, and Jan Peters (2019). "Deep Lagrangian Networks: Using Physics as Model Prior for Deep Learning." In: *International Conference on Learning Representations (ICLR)*.

Markovsky, Ivan and Paolo Rapisarda (Dec. 2008). "Data-driven simulation and control." *International Journal of Control* 81.12, pp. 1946–1959.

Misyris, George S, Andreas Venzke, and Spyros Chatzivasileiadis (2020). "Physics-informed neural networks for power systems." In: *2020 IEEE Power & Energy Society General Meeting (PESGM)*.

Pfaff, Tobias, Meire Fortunato, Alvaro Sanchez-Gonzalez, and Peter W Battaglia (2020). "Learning Mesh-Based Simulation with Graph Networks." *arXiv preprint arXiv:2010.03409*.

Poli, Michael, Stefano Massaroli, Junyoung Park, Atsushi Yamashita, Hajime Asama, and Jinkyoo Park (2019). "Graph neural ordinary differential equations." *arXiv preprint arXiv:1911.07532*.

Rackauckas, Chris, Ranjan Anantharaman, Alan Edelman, Shashi Gowda, Maja Gwozdz, Anand Jain, Chris Laughman, Yingbo Ma, Francesco Martinuzzi, Avik Pal, et al. (2021). "Composing Modeling and Simulation with Machine Learning in Julia." *arXiv preprint arXiv:2105.05946*.

Rackauckas, Christopher and Qing Nie (2017). "DifferentialEquations.jl A Performant and Feature-Rich Ecosystem for Solving Differential Equations in Julia." *The Journal of Open Research Software* 5.1.

Raissi, Maziar, Paris Perdikaris, and George E Karniadakis (2019). "Physics-informed neural networks: A deep learning framework for solving forward and inverse problems involving nonlinear partial differential equations." *Journal of Computational Physics* 378, pp. 686–707.

Rodrigues, Francisco A., Thomas K. DM. Peron, Peng Ji, and Jürgen Kurths (2016). "The Kuramoto model in complex networks." *Physics Reports* 610. The Kuramoto model in complex networks, pp. 1–98. ISSN: 0370-1573.

Sanchez-Gonzalez, Alvaro, Jonathan Godwin, Tobias Pfaff, Rex Ying, Jure Leskovec, and Peter Battaglia (2020). "Learning to simulate complex physics with graph networks." In: *International Conference on Machine Learning (ICML)*.

Satorras, Victor Garcia, Zeynep Akata, and Max Welling (2019). "Combining generative and discriminative models for hybrid inference." *arXiv preprint arXiv:1906.02547*.

Scarselli, Franco, Marco Gori, Ah Chung Tsoi, Markus Hagenbuchner, and Gabriele Monfardini (2008). "The graph neural network model." *IEEE transactions on neural networks* 20.1, pp. 61–80.

Tzen, Belinda and Maxim Raginsky (2019). "Neural stochastic differential equations: Deep latent gaussian models in the diffusion limit." *arXiv preprint arXiv:1905.09883*.

Watts, Duncan J and Steven H Strogatz (1998). "Collective dynamics of small-worldnetworks." *Nature* 393.6684, pp. 440–442.

Zang, Chengxi and Fei Wang (2020). "Neural dynamics on complex networks." In: *Proceedings of the 26th ACM SIGKDD International Conference on Knowledge Discovery & Data Mining*.

# Appendices

## A. Experiment Details

Source code for the training pipeline, tasks, and models used in this work, is available as part of the supplementary material.

We used the same *Adam* Kingma et al., 2015 optimizer for all our experiments and a learning rate of 0.001, and a batch size of 128. For solving the differential equations both during ground truth data generation as well as with the neural ODEs, we use the Tsitouras 5/4 Runge-Kutta (Tsit5) method from DifferentialEquations.jl (Christopher Rackauckas et al., 2017).

### A.1. Coupled Pendulum

The coupled pendulum dynamics are defined as

$$\ddot{\theta}_1 = \frac{\sin\theta_1 * (m_1 l_1 \dot{\theta}_1^2 - g - kl_1) + kl_2\sin\theta_2}{m_1 l_1 \cos\theta_1} \qquad (6)$$

$$\ddot{\theta}_2 = \frac{\sin\theta_2 * (m_2 l_2 \dot{\theta}_2^2 - g - kl_2) + kl_1\sin\theta_1}{m_2 l_2 \cos\theta_2}$$

Where $\theta_i, \dot{\theta}_i$ refer to the angle and the angular velocity of the $i^{\text{th}}$ pendulum respectively. $m_i$ and $l_i$ are the mass and length corresponding to the $i^{\text{th}}$ pendulum, which are chosen to be 1.0kg and 1.5m respectively. $k$ is the spring constant of the coupling string (chosen to be 2.0), and $g$ is the gravitational acceleration. The system state is defined as $[\theta_1, \dot{\theta}_1, \theta_2, \dot{\theta}_2]$.

We train the MP-NODE on a dataset of 500 trajectories, each randomly initialized with state values between $[-\pi/2, \pi/2]$ for the $\theta$ and $[-1, 1]$ for $\dot{\theta}$, with a time step of 0.1s and each trajectory 10s long. The dataset is normalized through Z-score normalization. We use the mean squared error loss during training.

### A.2. Lorenz Attractor Systems

A Lorenz system is a simplified model of atmospheric convection, described through a 3-dimensional system that exhibits chaotic behavior in specific cases. The system has a three-dimensional state whose evolution is described as $\dot{x} = \sigma(y-x); \dot{y} = x(\rho-z) - y; \dot{z} = xy - \beta z$. In our implementation, we use the values of $\sigma = 10.0, \rho = 28.0, \beta = 2.666$ for the Lorenz attractor which match the values originally used in Lorenz, 1963. We consider multiple Lorenz systems coupled in a fully connected manner.

We generate data from multiple Lorenz attractor configurations, the variables being the number of Lorenz nodes (3, 7, 10), the length of the trajectories (2.5s, 5s) and the coupling magnitude. We generate the coupling matrix with random coupling strengths between 0 and 1, with the diagonal elements set to zero (as the nodes are not connected to themselves). We use an additional coupling magnitude parameter that the coupling matrix is multiplied with to strengthen/weaken the coupling effect: which is set to 0.01 for low coupling and 1.0 for high coupling.

We train our MP-NODE primarily on the Lorenz3 system, with trajectories that are 2.5s long with a dt (timestep) of 0.05s and low coupling magnitude. We perform Z-score normalization on the data and use Huber loss along the time dimension as the training loss function. This model is later finetuned on the other configurations.

### A.3. Gene Dynamics

The gene regulatory dynamics for a grid-based system are governed by the Michaelis-Menten equation that is shown below (Alon, 2019).

$$\frac{dx_i(t)}{dt} = -b_i x_i{}^g + \sum_{j=1}^n A_{ij}\frac{x_j{}^h}{x_j{}^h + 1} \qquad (7)$$

We solve the Michaelis-Menten equation over two sizes of 2D grids: $4 \times 4$ (gene_small) and $8 \times 8$ (gene_large). The adjacency matrices were generated randomly according to three network topologies:

a) Erdós-Rényi (ER) (Erds et al., 1960); b) Barabási-Albert (BA) (Barabási et al., 1999) and c) Wattz-Strogatz (WS) (Watts et al., 1998). For each grid with $n$ cells in total, each cell is connected to $n/2$ other cells. The initial state at each cell of the grid was chosen randomly to be between 0 and 50. Our implementation loosely follows that of Zang et al., 2020.

The training dataset contains 200 trajectories from gene_small with five different adjacency matrices of the power-law network type, of which 70% are used for training and the rest for evaluation. Each trajectory is 5s long with a timestep of 0.1s. For training the MP-NODE on this system, we use the mean squared error loss. This model is later finetuned on gene_large, and the other network topologies.

### A.4. Kuramoto Systems

The Kuramoto model (Kuramoto, 2003) describes the behavior of large sets of coupled oscillators. Variations of the Kuramoto model find applications in a variety of fields, such as neuroscience (Rodrigues et al., 2016), power systems (Guo et al., 2021) and vehicle coordination (Dörfler et al., 2014). The dynamics are defined as:

$$\frac{dx_i(t)}{dt} = b_i + \sum_{j=1}^{N} A_{ij} \sin(x_j - x_i) \tag{8}$$

We also simulate the Kuramoto systems with adjacency matrices according to three network topologies: random, power-law and small-world. We generate data for Kuramoto systems with 10 nodes, with each node connected to 5 other nodes. The initial state at each cell of the grid was chosen randomly to be between -1 and 1. The timestep for the trajectories was set to 0.05s.

We train the main MP-NODE model on a dataset of 500 trajectories, of which 70% are used for training and the rest for evaluation. We perform Z-score normalization on the data and use Huber loss along the time dimension as the training loss function. This model is later finetuned on data from other network topologies.

### A.5. Load Carrying Quadrotors

Quadrotor dynamics are defined by:

$$\dot{x} = \begin{bmatrix} \dot{r} \\ \dot{q} \\ \dot{v} \\ \dot{\omega} \end{bmatrix} = \begin{bmatrix} v \\ \frac{1}{2} q \otimes \hat{\omega} \\ g + \frac{1}{m} \left( R(q) F(u) + F_c(u_5, x, x^\ell) \right) \\ J^{-1} \left( \tau(u) - \omega \times J\omega \right) \end{bmatrix} \tag{9}$$

where $r \in \mathbb{R}^3$ is the position, $q$ is the unit quaternion, $R(q) \in \mathbb{SO}(3)$ is quaternion dependent rotation matrix from body frame to world frame, $v \in \mathbb{R}^3$ is the linear velocity in the world frame, $\omega \in \mathbb{R}^3$ is the angular velocity in the body frame. $g$ is the gravity vector, $m$ is the mass of the individual quadrotor, $J \in \mathbb{S}^3$ is the moment of inertia tensor, $q_2 \otimes q_1$ denotes quaternion multiplication, and $\hat{\omega}$ denotes a quaternion with zero scalar part and $\omega$ vector part. The forces ($F \in \mathbb{R}^3$) and torques ($\tau \in \mathbb{R}^3$) are in the body frame. For our experiments, the coupling between the quadrotors was provided via (unobserved) state vector of the load $x^\ell \in \mathbb{R}^6$. Therefore the state vector $x \in \mathbb{R}^{13}$ and the control vector $u \in \mathbb{R}^5$. For more details please refer to (Jackson et al., 2020).

The training dataset was generated using the batch trajectory optimizer in Jackson et al., 2020. We generated a total of 463 trajectories for the 3 quadrotor system with different initial conditions of the load and the quadrotors. We used 70% trajectories for full training and rest for evaluation. The trajectories were 10s long with timestep, $dt = 0.2$s. Similar method was used to generate data for 6 quadrotor system. The data was standardized using the Z-score transform. Mean-Square-Error between the predicted trajectory and the ground truth trajectory was used as the training objective.

# B. Additional Experiments

## B.1. Coupled Pendulum

We show coupled pendulum dynamics evolution and the corresponding messages from a different initial state in Figure 10.

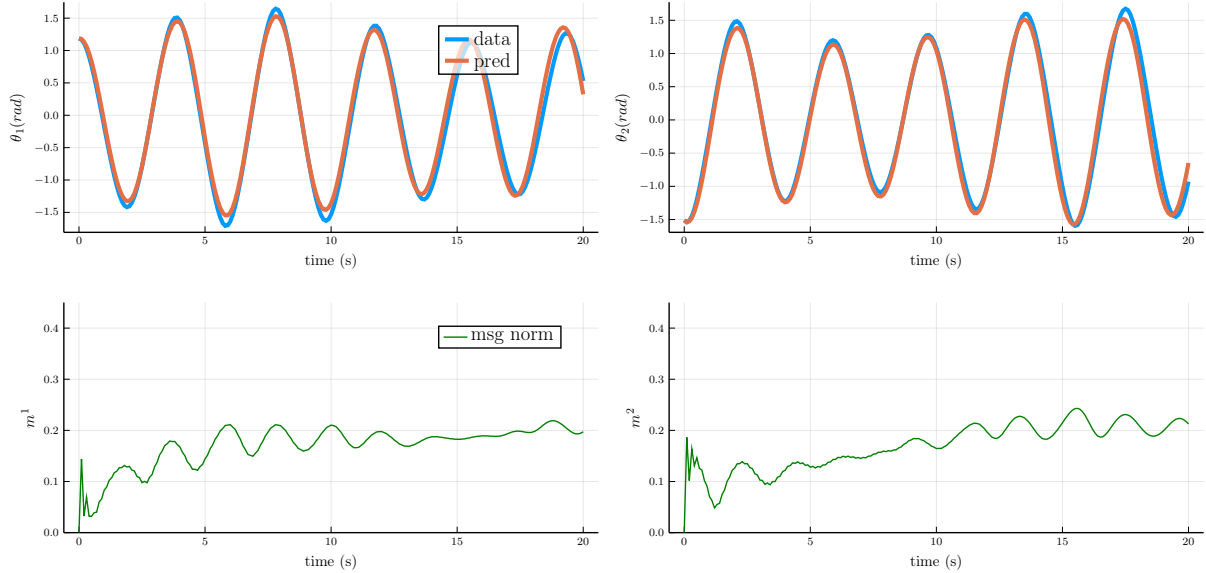

**Figure 10:** Additional example of the state of each pendulum and corresponding messages that allow for accurate predictions.

## B.2. Lorenz Systems

In section 4.3, we discuss the zero-shot generalization ability of the MP-NODE. In Figure 11a and Figure 11b, we show results of a MP-NODE model trained only on Lorenz3, but tested on Lorenz7 and Lorenz10 without any finetuning. We observe reasonably good performance on Lorenz7 and 10 which are unseen configurations during training.

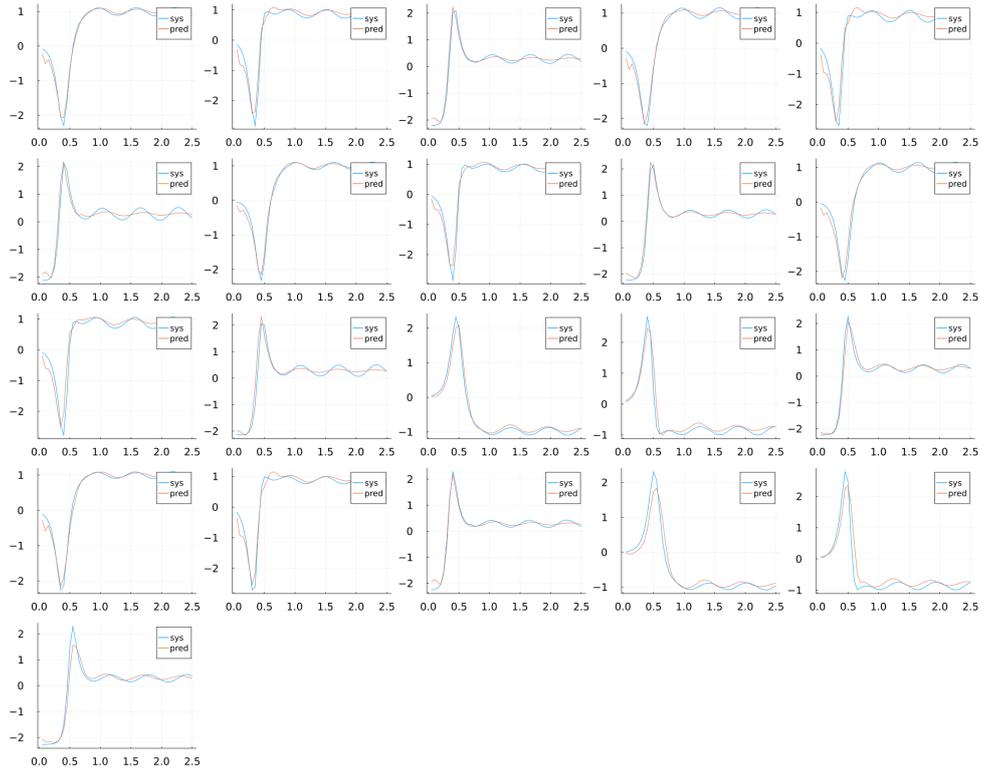

**(a)** Lorenz3 trained MP-NODE evaluated on Lorenz7.

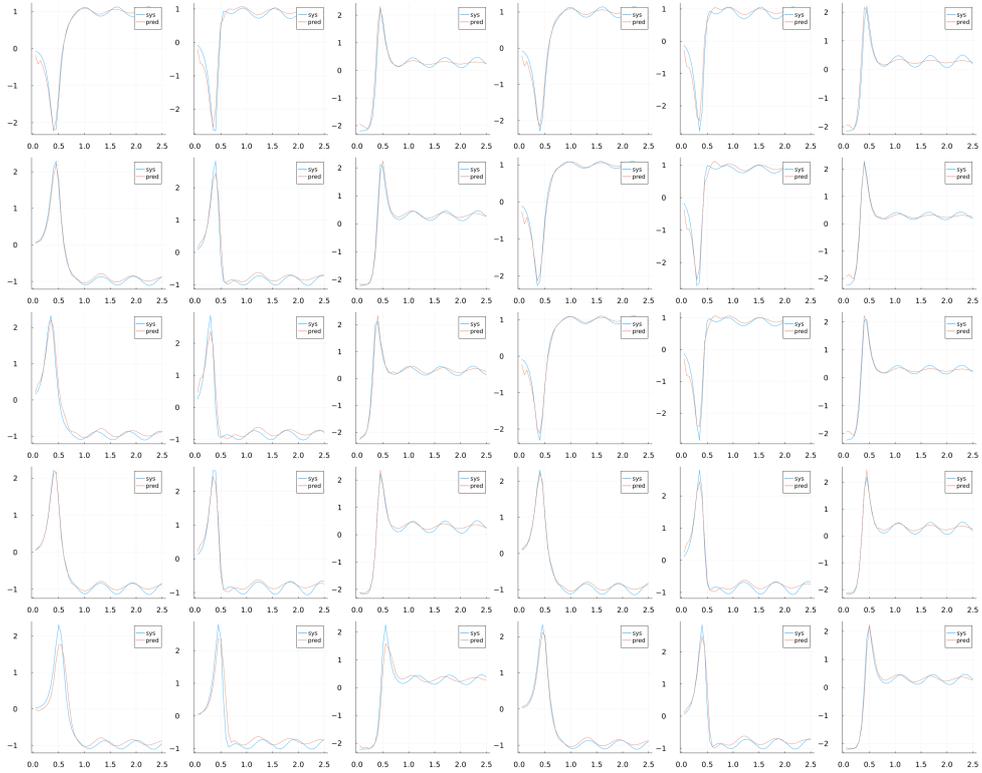

**(b)** Lorenz3 trained MP-NODE evaluated on Lorenz10.

**Figure 11:** Zero-shot generalization of MP-NODE on Lorenz systems. From left to right, top to bottom - plots depict the predicted vs. ground truth dynamics on X, Y, Z states of all 7 or 10 nodes.

Learning Modular Simulations for Homogeneous Systems

### B.3. Quadrotor Systems

With increasing dimensions of messages, it is difficult to analyze the roles they play. To this end we apply Principal Component Analysis (PCA) for all 39 messages over time for 100 trajectories and visualize the evolution of first three principal components over the time horizon of the trajectory in Figure 12. There is an overall trend of increasing values of these components over time. We believe that these increasing values hint at the messages acting in to counteract the problem of simulation drift over time. The pattern of smaller values at some time indices likely corresponds to the nature of dynamic interaction when the quadrotors reconfigure themselves to carry the load through the narrow doorway.

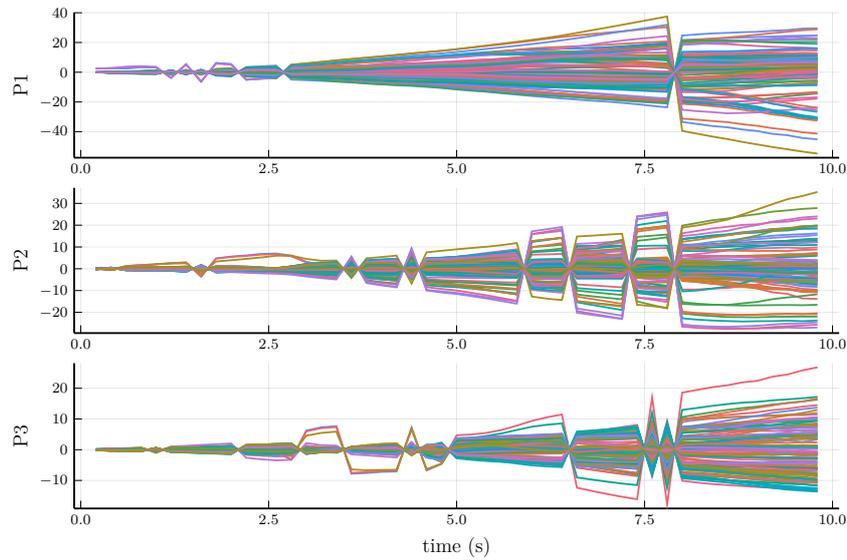

**Figure 12:** First three principal components of messages over time for 100 different trajectories.